\pdfoutput=1

\documentclass[11pt]{article}

\usepackage{acl}

\usepackage{times}
\usepackage{latexsym}

\usepackage[T1]{fontenc}

\usepackage[utf8]{inputenc}

\usepackage{microtype}

\usepackage{inconsolata}

\usepackage{graphicx}

\title{Holistic Evaluation for Interleaved Text-and-Image Generation}

\author{Minqian Liu$^{\spadesuit}$ \quad Zhiyang Xu$^{\spadesuit}$ \quad Zihao Lin$^{\spadesuit}$ \quad Trevor Ashby$^{\spadesuit}$ \\ \textbf{Joy Rimchala}$^\heartsuit$ \quad
\textbf{Jiaxin Zhang}$^\heartsuit$ \quad \textbf{Lifu Huang}$^{\spadesuit,\clubsuit}$
\\
  $^{\spadesuit}$Virginia Tech \quad  
$^\heartsuit$Intuit AI Research \quad $^\clubsuit$University of California, Davis \\ 
  {\tt \{minqianliu, zhiyangx, zihaol, trevorashby, lifuh\}@vt.edu } \\ 
  {\tt \{joy\_rimchala, jiaxin\_zhang\}@intuit.com } \\
  }

\usepackage{graphicx}               
\usepackage{tabularx}               
\usepackage{booktabs}
\usepackage{amsmath}
\usepackage{stmaryrd}
\usepackage{xcolor}
\usepackage{soul}
\usepackage[linesnumbered,ruled,vlined]{algorithm2e}

\newcolumntype{C}{>{\centering\arraybackslash}X}
\usepackage{multirow}               
\usepackage{diagbox}                
\usepackage{hhline}                 
\usepackage{color, colortbl}                  
\usepackage{amsmath}                
\usepackage{amssymb}                
\usepackage{pifont}
\usepackage{mathtools}              
\usepackage{enumitem}               

\usepackage{subfigure}
\usepackage{booktabs}
\usepackage{extarrows}
\usepackage{makecell}

\definecolor{lightblue}{rgb}{0.0, 0.5, 1.0}

\definecolor{RoseQuartzBg}{HTML}{F7CAC9}
\definecolor{RoseQuartz}{HTML}{F5A798}
\definecolor{Serenity}{HTML}{92A8D1}
\definecolor{OrangeRed}{rgb}{1.0, 0.27, 0.0}
\definecolor{Red}{rgb}{1.0, 0.0, 0.0}
\definecolor{Turquoise}{HTML}{0F4C81}
\usepackage{xparse}
\NewDocumentCommand{\lifu}{ mO{} }{\textcolor{OrangeRed}{\textsuperscript{\textit{Lifu}}\textsf{\textbf{\small[#1]}}}}
\NewDocumentCommand{\mo}{ mO{} }{\textcolor{blue}{\textsuperscript{\textit{Mo}}\textsf{\textbf{\small[#1]}}}}
\NewDocumentCommand{\minqian}{ mO{} }{\textcolor{lightblue}{\textsuperscript{\textit{Minqian}}\textsf{\textbf{\small[#1]}}}}
\NewDocumentCommand{\zhiyang}{ mO{} }{\textcolor{blue}{\textsuperscript{\textit{Zhiyang}}\textsf{\textbf{\small[#1]}}}}
\NewDocumentCommand{\sijia}{ mO{} }{\textcolor{Red}{\textsuperscript{\textit{Sijiia}}\textsf{\textbf{\small[#1]}}}}
\NewDocumentCommand{\ying}{ mO{} }{\textcolor{teal}{\textsuperscript{\textit{Ying}}\textsf{\textbf{\small[#1]}}}}
\NewDocumentCommand{\zihao}{ mO{} }{\textcolor{orange}{\textsuperscript{\textit{zihao}}\textsf{\textbf{\small[#1]}}}}

\newcommand{\cmark}{\ding{51}}%
\newcommand{\xmark}{\ding{55}}%

\newcommand{\dataset}{\textsc{InterleavedBench}}

\newcommand{\eval}{\textsc{InterleavedEval}}

\begin{document}

\vspace{-5pt}
\twocolumn[{%
    \renewcommand\twocolumn[1][]{#1}
    \maketitle
    \centering
    \vspace{-30pt}
    \url{https://vt-nlp.github.io/InterleavedEval/}
    \vspace{1em}
    \begin{center}
        \centering
        \includegraphics[width=0.9\textwidth]{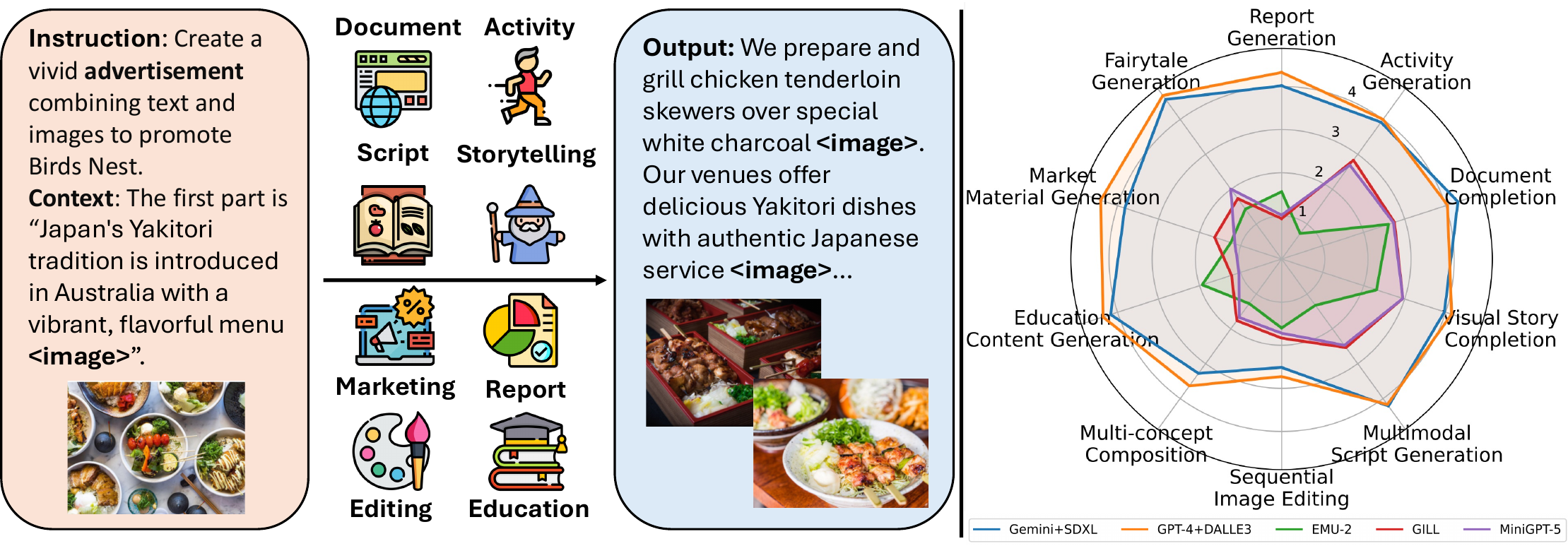}
        \captionof{figure}{Overview of our \dataset{}, a comprehensive benchmark that covers 10 diverse use cases for interleaved text-and-image generation, and the evaluation results of \eval{} based on GPT-4o. }
        \vspace{2ex}
        \label{fig:teaser}
    \end{center}
}]


\begin{abstract}

Interleaved text-and-image generation has been an intriguing research direction, where the models are required to generate both images and text pieces in an arbitrary order. Despite the emerging advancements in interleaved generation, the progress in its evaluation still significantly lags behind. Existing evaluation benchmarks 
do not support arbitrarily interleaved images and text for both inputs and outputs, and they only cover a limited number of domains and use cases. Also, current works predominantly use similarity-based metrics which fall short in assessing the quality in open-ended scenarios. To this end, we introduce \dataset{}, the first benchmark carefully curated for the evaluation of interleaved text-and-image generation. \dataset{} features a rich array of tasks to cover diverse real-world use cases. In addition, we present \eval{}, a strong reference-free metric powered by GPT-4o to deliver accurate and explainable evaluation. We carefully define five essential evaluation aspects for \eval{}, including \textit{text quality}, \textit{perceptual quality}, \textit{image coherence}, \textit{text-image coherence}, and \textit{helpfulness}, to ensure a comprehensive and fine-grained assessment. Through extensive experiments and rigorous human evaluation, we show that our benchmark and metric can effectively evaluate the existing models with a strong correlation with human judgments surpassing previous reference-based metrics. We also provide substantial findings and insights to foster future research in interleaved generation and its evaluation.\footnote{The source code and datasets are publicly available at \url{https://huggingface.co/mqliu/InterleavedBench} for research purposes.}

\end{abstract}

\section{Introduction}

\begin{figure*}[htbp]
	\centering
	\includegraphics[width=\linewidth, trim={0 0 0 0},clip ]{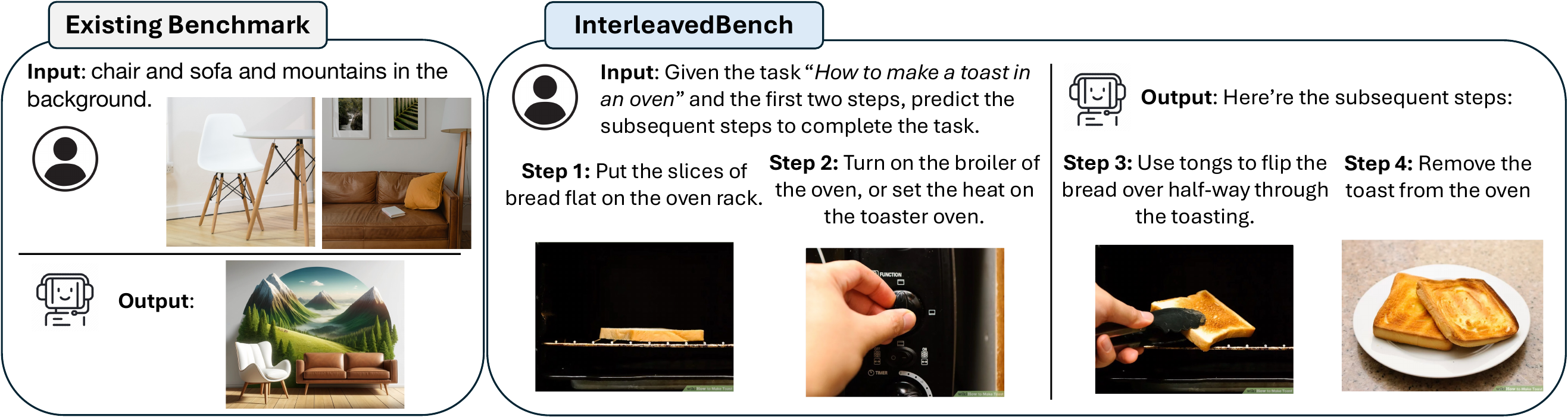}
	\caption{Comparison between the existing benchmark (multi-concept image composition~\cite{kumari2022customdiffusion}) and our \dataset{}. Compared with the existing benchmark, \dataset{} has the following features: (1) both input and output can have arbitrarily interleaved text and images, and (2) each instance has a detailed instruction to benchmark models' instruction-following capability.
 }
\label{fig_data_comp}
\end{figure*}

Multimodal learning has been a rapidly developing research field given the recent advancements in Large Multimodal Models (LMMs)~\cite{xu2023multiinstruct,Dai2023InstructBLIPTG,liu2023llava}.
While these models can perform diverse tasks such as detailed image description and visual question answering, the outputs are limited to the text-only format, which hinders their broader applications. More recently, there has been a growing focus on enhancing LMMs with the capability of \textit{interleaved generation}, i.e., generating multimodal content that seamlessly integrates both text and one or multiple images~\cite{gill,dreamllm,emu,emu2}. This opens new avenues for applications in diverse challenging scenarios, such as creative content generation \cite{anantrasirichai2022artificial}, visual storytelling~\cite{vist, lukin2018pipeline}, and multimodal script generation~\cite{wikihow,multiscript}.


While the LMMs for interleaved generation are continuously gaining stronger capabilities, progress in the \textit{evaluation} of interleaved generation significantly lags behind with several critical challenges remaining. \textbf{First}, most existing works for interleaved generation quantitatively benchmark the models on text-to-image tasks where the output is usually one single image~\cite{gill,dreamllm}. However, such evaluation methods would fail to assess model performance in the real-world scenarios of interleaved generation, where the output usually consists of interleaved text and images. \textbf{Second}, apart from human evaluation which is costly and time-consuming, existing works still heavily rely on reference-based metrics such as BLEU~\cite{bleu} FID~\cite{fid} that measure the similarity between generated samples and gold references. Such similarity-based metrics often fail to accurately capture outputs' quality, especially in open-ended tasks such as creative generation and visual storytelling. \textbf{Third}, the evaluation of interleaved generation is complex and involves many different aspects, such as \textit{perceptual quality}, \textit{coherence} between text and images, and \textit{helpfulness} of the overall content. One single aspect is usually insufficient to reflect the overall quality. For example, despite the images in one output having good perceptual quality, the output can still be not helpful to users if the generated content is not coherent with the context, e.g., the request from users.





To address these critical limitations, we introduce \dataset{}, the first benchmark for holistic evaluation of interleaved text-and-image generation. We construct \dataset{} with a high-quality and diverse collection of interleaved generation scenarios that encompass a wide range of real-world use cases, including creative generation, multimodal script generation, visual storytelling, and many others. We compare our \dataset{} and one existing benchmark~\cite{kumari2023multi} closest to our dataset in Figure~\ref{fig_data_comp}.
To support the evaluation, we also introduce \eval{}, a strong reference-free evaluation metric based on GPT-4o~\cite{OpenAI2024}, the current state-of-the-art LMM. \eval{} can take in any evaluation instructions and provide a fine-grained evaluation along with detailed explanations.
We carefully curate a multi-aspect evaluation criterion to ensure a holistic evaluation for \eval. 
Specifically, we define five essential aspects for interleaved evaluation, including \textit{text quality, perceptual quality, image coherence, text-image coherence}, and \textit{helpfulness}, following the principles that (1) these aspects are generally applicable in different scenarios, (2) these aspects are atomic and orthogonal to each other, and (3) the combination of these aspects can comprehensively cover the critical dimensions in interleaved generation.

Extensive experiments and rigorous human evaluation demonstrate that \textbf{(1)} Our curated \dataset{} posts unique and significant challenges to the existing integrated LMMs (e.g., GILL~\cite{gill} and EMU-2~\cite{emu2}) for interleaved generation, where the quality of their outputs are far from satisfying. The pipeline systems combined with a strong LMM (e.g., GPT-4o) and a separate image generation model (e.g., DALLE3~\cite{BetkerImprovingIG}) generally achieve better results but still struggle on certain tasks; \textbf{(2)} \eval{} can achieve a good correlation with human judgments with significant improvement over previous automatic evaluation metrics; \textbf{(3)} The evaluation of interleaved generation remains a very challenging direction due to its complexity and the limitation of the existing LMM-based evaluator. We believe that our work can provide useful resources and insights for interleaved generation and its evaluation.  



\section{Related Work}

\begin{figure*}[htbp]
	\centering
	\includegraphics[width=\linewidth, trim={0 0 0 0},clip ]{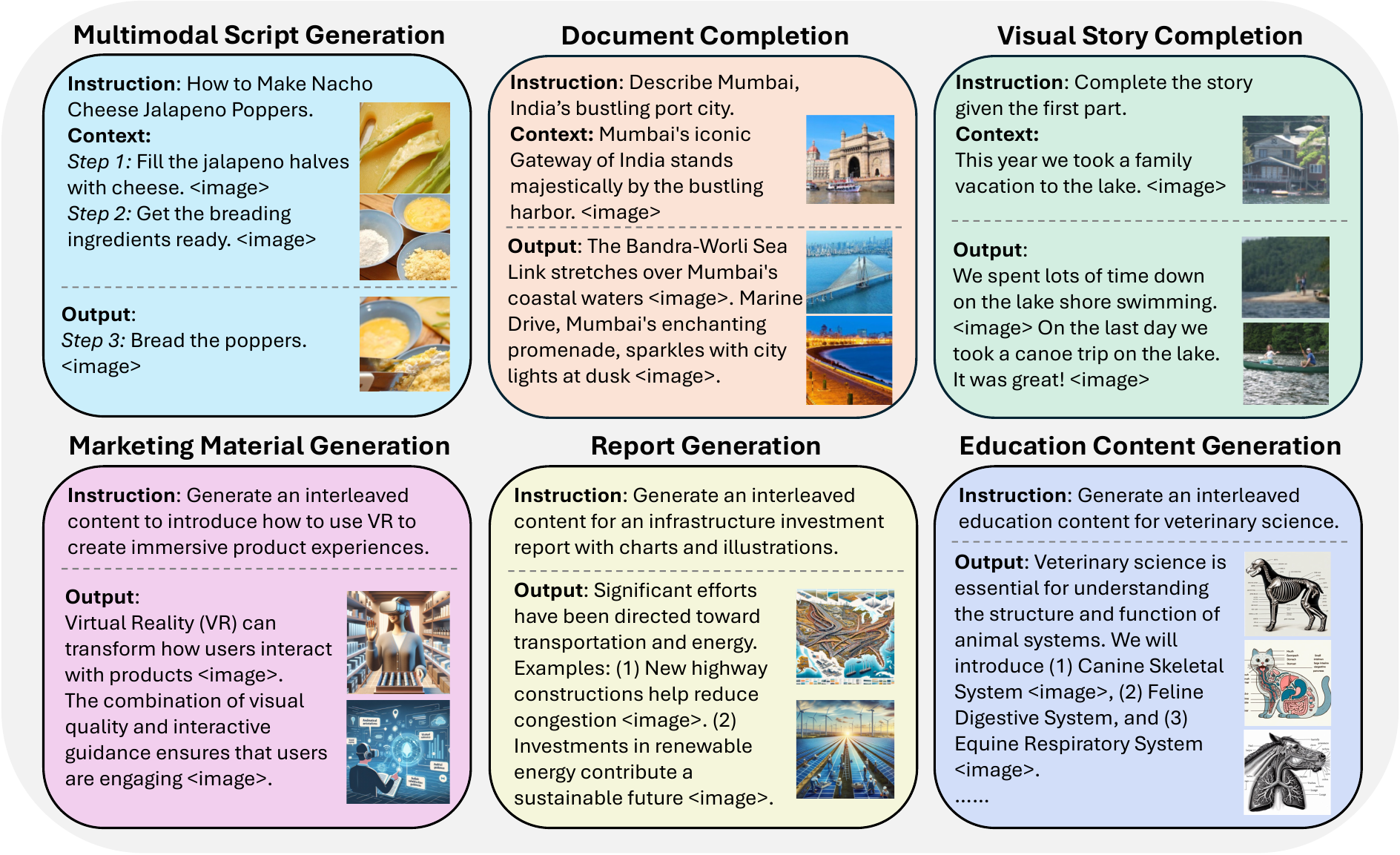}
	\caption{Illustration of examples in our \dataset{} from six representative use cases. }
\label{fig_main_usecase}
\end{figure*}

\paragraph{Large Multimodal Models for Interleaved Generation}

The advent of large multimodal models (LMMs)~\cite{gill, emu2} has significantly advanced the field of interleaved text-and-image generation. Previous models such as DALL-E~\cite{ramesh2021zero} and Stable Diffusion~\cite{sdxl} have demonstrated impressive capabilities in generating high-quality images conditioned on textual descriptions. However, previous focus has predominantly been on unidirectional generation tasks, either from text to image or image to text, without considering the interleaved generation scenarios where text and images are seamlessly integrated within the same output. Recent works have begun to address this gap, with the LMMs extended with diffusion models, exploring the generation of mixed text and image outputs~\cite{gill,emu,dreamllm,mmInterleave,anyGPT,llavaInteract}. These models leverage advanced architectures and training techniques to enhance their ability to produce coherent and contextually relevant interleaved content. Despite these advancements, the evaluation of such models remains an underexplored area, with most evaluations still relying on separate assessments of text and image quality or simplistic reference-based metrics. Our proposed \dataset{} benchmark aims to bridge this gap by providing a holistic evaluation framework tailored specifically for interleaved text-and-image generation.

\paragraph{Evaluation of Multimodal Generation Tasks}
Evaluating multimodal generation tasks presents unique challenges due to the inherent complexity of assessing both textual and visual components simultaneously. Existing metrics for text generation, such as BLEU~\cite{bleu}, ROUGE~\cite{rouge}, and LLM-based evaluators~\cite{unieval,geval,xeval}, fall short when applied to multimodal outputs as they fail to capture the visual quality and coherence with textual content. Similarly, visual generation metrics like FID~\cite{fid} and IS~\cite{inception_score} are inadequate for evaluating the textual elements accompanying the images. To address this, recent studies have employed multimodal metrics~\cite{gpt4veval,viescore}, such as CLIPScore~\cite{clipscore}, which leverages the alignment capabilities of the CLIP model to measure the similarity between generated images and their corresponding textual descriptions. 
However, CLIPScore can only measure the alignment between text and images, which is not sufficient to evaluate the quality of generated output comprehensively.
Moreover, human evaluations, although more reliable, are resource-intensive and cannot be scalable.
In terms of evaluation benchmarks in multimodal learning, existing works mostly focus on evaluating the tasks with single-modality output~\cite{blink,seedbench,wildvision,muribench,fu2024isobench}, such as conditional text-to-image generation~\cite{chen2024subject,ku2024imagenhub}, where the primary focus is solely the quality of generated images.
Our \dataset{} benchmark introduces a novel approach to evaluate interleaved text-and-image generation by incorporating multiple aspects of quality assessment, thus providing a more nuanced and holistic evaluation framework.


\section{\dataset{}}







We introduce \dataset{}, the first comprehensive benchmark meticulously constructed to evaluate text-and-image interleaved generation. Figure~\ref{fig_main_usecase} shows some examples from \dataset{}.


\subsection{Dataset Curation Process}

Our dataset includes two subsets: a \textbf{context-based} subset where the instances contain a multimodal context of interleaved text and images in the input (first row in Figure~\ref{fig_main_usecase}), and a \textbf{context-free} subset with text-only inputs (second row in Figure~\ref{fig_main_usecase}). The context-free subset can assess whether the model can creatively generate interleaved content based on the text-only instruction, while the context-based subset can better benchmark the coherence and consistency of generated outputs.

\paragraph{Collection of Context-based Subset}
\textbf{Firstly}, we collect the source data of the context-based subset from existing academic datasets or web resources. Specifically, we collect the data of multimodal script generation from WikiHow~\cite{wikihow}, visual story completion from VIST~\cite{vist}, activity generation from the dense captions and the extracted video frames in ActivityNet Captions~\cite{activitynet}, sequential image editing from MagicBrush~\cite{MagicBrush}, and multi-concept image composition from CustomDiffusion~\cite{kumari2022customdiffusion}. For web resources, we apply an automatic data filtering pipeline to discard the samples with poor quality to obtain a small set of source data. We detail our data filtering pipeline in Appendix~\ref{app:dataset}.
\textbf{Secondly}, after collecting the source data (either from academic benchmarks or web resources), we then apply a human selection process to manually select the samples based on data quality and diversity (i.e., avoiding selecting similar samples).
\textbf{Finally}, we ask human experts to annotate an instruction $I$ for each sample based on the collected content.
We include the details of the data selection and instruction annotation process in Appendix~\ref{app:dataset}.
For the samples that are originally interleaved articles, we pick the first $k$ images and their associated text as the \textit{context} $\mathcal{C}$ for the input. $k$ is randomly sampled for each example and ranges from 1 to the maximum number of images minus 1 since we need to ensure the output contains at least one image. The rest of the images and text are used as the gold reference. 

\begin{table*}[h]
\center 
\resizebox{0.9\textwidth}{!}
{
\begin{tabular}{l|c|c|c|c}
\toprule
\textbf{Dataset Name} & \textbf{Detailed Instruction} & \textbf{Image Input} & \textbf{Text Output} & \textbf{Image Output} \\ 

\midrule
MagicBrush~\cite{MagicBrush} & No & Single & No & Single  \\
DreamBench~\cite{chen2024subject} & No & Multiple & No & Single \\
CustomDiffusion~\cite{kumari2022customdiffusion} & No & Multiple & No & Single  \\
DreamEditBench~\cite{li2023dreamedit} & No & Multiple & No & Single \\
Mantis-Eval~\cite{mantis} & Yes & Multiple & Yes & No \\
BLINK~\cite{blink} & Yes & Multiple & Yes & No \\
MuriBench~\cite{muribench} & Yes & Multiple & Yes & No \\
\midrule
\dataset{} (Ours) & Yes & Multiple & Yes & Multiple  \\
\bottomrule
\end{tabular}
}
\caption{Comparisons between \dataset{} and existing open-sourced multimodal evaluation benchmarks. The highlighted features of our benchmark include detailed instructions and multiple images in input and/or output that are arbitrarily interleaved with text.}
\label{tab:data_comparison}
\vspace{-3mm}
\end{table*}

\begin{figure}[tbp]
	\centering
	\includegraphics[width=\linewidth, trim={2cm 0.8cm 1cm 0.8cm},clip ]{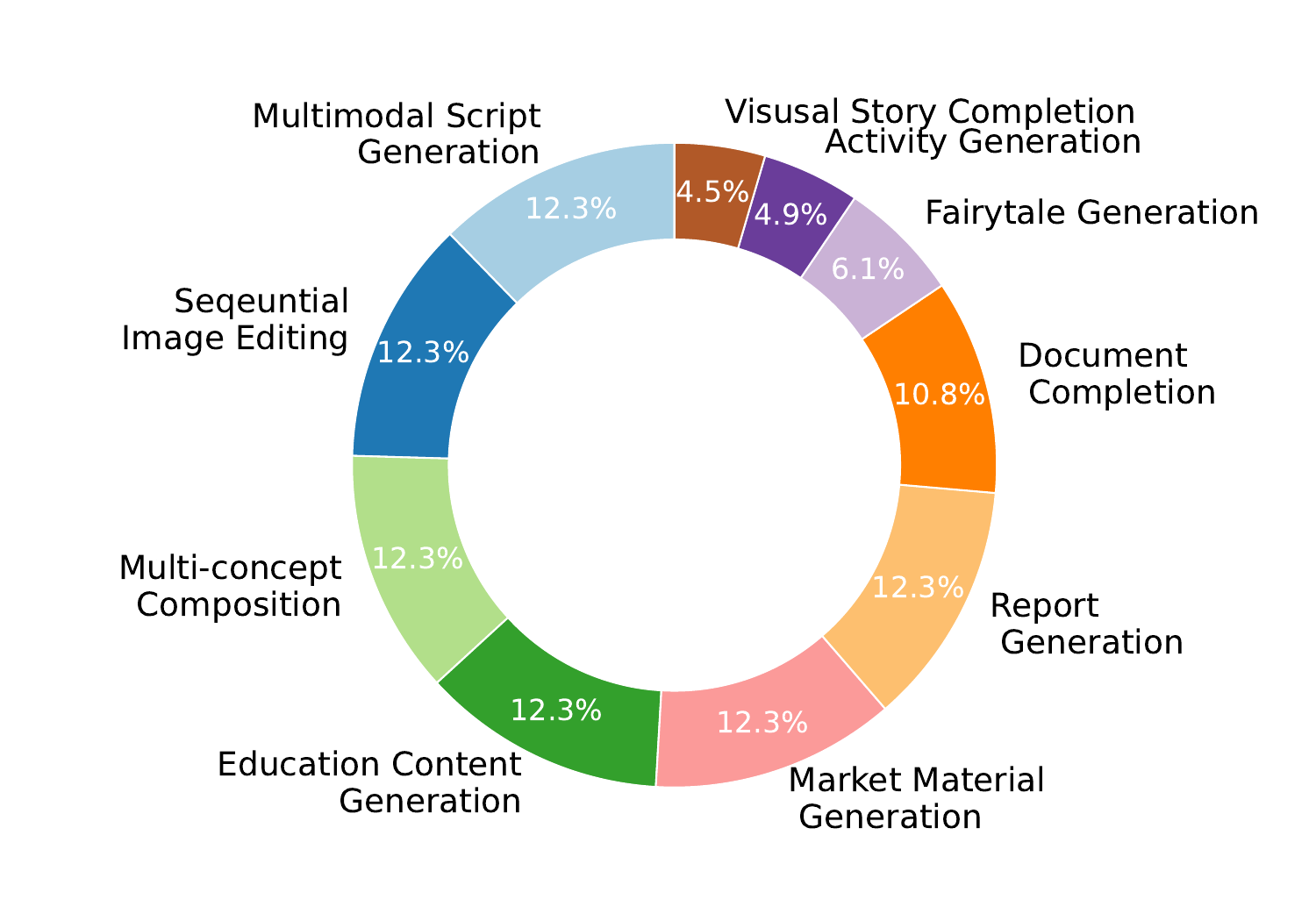}
	\caption{The distribution of the use cases in \dataset{}.
 }
\label{fig_distribution}
 \vspace{-3mm}
\end{figure}

\paragraph{Collection of Context-free Subset}
The context-free subset consists of the use cases of \textit{marketing material generation, report generation,} \textit{education content generation}, and \textit{fairytale generation} as they are common and practical scenarios for interleaved generation.
We first leverage GPT-4o to generate a set of instances for each use case. For example, in marketing material generation, one instance is ``\textit{creating marketing campaigns around holidays to boost sales}''. Then, we use GPT-4o to extend each instance into a more detailed instruction, e.g., ``\textit{Create an interleaved content that combines engaging text and eye-catching images for marketing campaigns around holidays to boost sales. Begin by researching holiday themes relevant to your products...}''. 
Finally, we ask human annotators to verify whether the instructions are reasonable and of good quality. Note that we do not have gold references in this subset.

\paragraph{Dataset Statistics}
In total, we finally collect 815 instances across 10 use cases, including \textit{multimodal script generation}, \textit{document completion}, \textit{visual story completion}, \textit{marketing material generation}, \textit{report generation}, \textit{education content generation}, \textit{activity generation}, \textit{sequential image editing}, and \textit{multi-concept image composition}. The detailed distribution of the use cases is shown in Figure~\ref{fig_distribution}. 

\subsection{Comparison with Existing Benchmark} We highlight the following key differences and unique challenges introduced by our \dataset{} compared with the existing benchmark. \textbf{(1): Output modality:} our benchmark requires the models to generate interleaved text and multiple images that could present in an arbitrary order, whereas exiting benchmarks~\cite{kumari2023multi} only cover the output with single modality or a single image (as shown in Figure~\ref{fig_data_comp}); \textbf{(2) Requirement on coherence:} given that both inputs and outputs in our benchmark can contain multiple pieces of text and images, our dataset can assess whether the outputs are coherent and consistent with input instruction and context, and within the outputs themselves; \textbf{(3) Instruction following:} Most existing conditional image generation datasets only contain simple instructions such as ``\textit{add a cat next to the person}''. On the contrary, each instance in our benchmark contains a detailed human-annotated instruction to describe the task. Thus, our dataset can evaluate models' instruction-following and generalization capabilities. We show the difference between our benchmark and existing datasets in Table~\ref{tab:data_comparison}.


\section{\eval{}}



In many use cases of interleaved generation, such as ``\textit{generate a story about Snow White using both text and images}'', comparing the output against a gold reference is unrealistic since the generation can be fairly open-ended. However, existing approaches predominantly use reference-based metrics, e.g., BLEU~\cite{bleu} and FID~\cite{fid}, to measure the quality of text and image, respectively. They usually fail to assess the quality accurately. 

To bridge the gap between existing metrics and the demand in more diverse and realistic scenarios, we present \eval{}, a strong reference-free metric based on GPT-4o, the current state-of-the-art LMM that supports arbitrarily interleaved inputs.
To obtain a holistic and comprehensive evaluation of interleaved generation, we define five fine-grained evaluation aspects, including \textit{text quality}, \textit{perceptual quality}, \textit{image coherence}, \textit{text-image coherence} and \textit{helpfulness}, and evaluate the output of each aspect separately. 
We show the detailed definition for each evaluation aspect in Table~\ref{tab:asp_def} in Appendix~\ref{app:evaluation}.
For each instance to be evaluated, the input of the evaluator consists of an \textbf{instruction} $I$ that indicates what should be accomplished, \textbf{system output} $X=(T_O, \mathcal{P}_O)$, where $T_O$ is the output text and $\mathcal{P}_O$ is the set of output images, the evaluation aspect $a$, and optionally, the \textbf{context} $\mathcal{C}$ of the task (e.g., the given text and images in models' inputs). 

We formulate the evaluation metric \eval{} as follows: 
We instruct the GPT-4o evaluator to output discrete scores from \{0, 1, 2, 3, 4, 5\} based on the detailed criteria shown in Table~\ref{tab:asp_def}, where 1 indicates the worst quality, 5 indicates the best quality, and 0 indicates output text and/or images are empty. We also instruct GPT-4o to provide a detailed explanation to improve the interpretability. 
Note that when the output text is empty, the scores on text-related aspects (\textit{text quality} and \textit{text-image quality}) are 0. Similarly, when the output image is empty, the scores on image-related aspects (\textit{perceptual quality}, \textit{image coherence}, and \textit{text-image quality}) are 0. Moreover, we do not apply the text-related aspects in sequential editing and subject-driven generation since the primary focus of these tasks is whether the image is generated correctly according to the instructions.





\section{Experiments}

\begin{table*}[t]
\center 
\resizebox{0.95\textwidth}{!}
{
\begin{tabular}{l|c|c|c|c|c|c}
\toprule
\textbf{Model} & \textbf{Text Quality} & \textbf{Perceptual Quality} & \textbf{Image Coherence} & \textbf{TIC} & \textbf{Helpfulness} & \textbf{AVG} \\ 

\midrule

MiniGPT-5 & 1.22 & 2.45 & 1.62 & 2.03 & 1.77 & 1.82 \\
GILL & 0.75 & 3.21 & 2.25 & 1.53 & 1.48 & 1.84 \\
EMU-2 & 1.26 & 2.28 & 1.89 & 1.34 & 1.64 & 1.68 \\
EMU-2 (Gold Text) & 1.56 & 3.35 & 2.89 & 1.43 & 2.10 & 2.27 \\
\midrule
Gemini1.5 + SDXL & \textbf{4.40} & 3.99 & \textbf{3.64} & 4.13 & 3.62 & 3.96 \\
GPT-4o + DALL·E 3 & 4.37 & \textbf{4.36} & 3.51 & \textbf{4.55} & \textbf{3.88} & \textbf{4.13} \\

\bottomrule
\end{tabular}
}
\caption{\textbf{Automatic evaluation} results of existing interleaved generation models on \dataset{} using \eval{} based on GPT-4o. TIC is the abbreviation for 'Text-Image Coherence'. The best results are highlighted in \textbf{bold}. }
\label{tab:main_result}
\end{table*}

\begin{table*}[t]
\center 
\resizebox{0.95\textwidth}{!}
{
\begin{tabular}{l|c|c|c|c|c|c}
\toprule
\textbf{Model} & \textbf{Text Quality} & \textbf{Perceptual Quality} & \textbf{Image Coherence} & \textbf{TIC} & \textbf{Helpfulness} & \textbf{AVG} \\ 

\midrule

GILL & 1.35 & 1.89 & 1.72 & 1.43 & 1.19 & 1.52 \\
EMU-2 & 1.23 & 1.74 & 1.87 & 1.24 & 1.2  & 1.46 \\
\midrule
Gemini1.5 + SDXL & \textbf{2.59} & 2.36 & \textbf{2.13} & \textbf{2.27} & 2.08 & 2.28 \\
GPT-4o + DALL·E 3 & 2.49 & \textbf{2.51} & 2.02 & 2.31 & \textbf{2.13} & \textbf{2.29} \\

\bottomrule
\end{tabular}
}
\caption{\textbf{Human evaluation} results of existing interleaved generation models on \dataset{}. TIC is the abbreviation for 'Text-Image Coherence'. The best results are highlighted in \textbf{bold}. 
Note that we use a scale of 0 to 3 for this evaluation, which is different from the scale used in Table~\ref{tab:main_result}.}
\label{tab:human_result}
\vspace{-3mm}
\end{table*}

\begin{table*}[t]
\center 
\resizebox{\textwidth}{!}
{
\begin{tabular}{l|c|c|c|c|c|c}
\toprule
\textbf{Metric} & \textbf{Ref-free?} & \textbf{Text Quality} & \textbf{Perceptual Quality} & \textbf{Image Coherence} & \textbf{TIC} & \textbf{Helpfulness}  \\ 

\midrule

BERTScore & \xmark & 0.21 & - & - & - & 0.37 \\
DreamSim & \xmark & - & 0.02 & 0.1 & - & 0.06 \\
Image-Image CLIPScore & \xmark & - & 0.08 & 0.2 & - & -0.01 \\
Text-Image CLIPScore & \cmark & - & - & - & 0.2 & 0.09 \\
\midrule
\eval{}-LLaVA & \cmark & 0.06 & \textbf{0.32} & 0.24 & 0.23 & 0.3 \\
\eval{}-GPT-4o & \cmark & \textbf{0.72} & 0.30 & \textbf{0.43} & \textbf{0.4} & \textbf{0.57} \\

\bottomrule
\end{tabular}
}
\caption{\textbf{Mete-evaluation on evaluation metrics} in terms of Spearman correlation between automatic evaluation results with human judgments. For baseline metrics, we only report the correlation on the corresponding aspects (e.g., BERTScore can correspond to \textit{text quality}) as well as \textit{helpfulness}. }
\label{tab:meta_eval}
\end{table*}


\subsection{Experiment Setup}

\paragraph{Baseline Models}
We benchmark the following baseline models which can be categorized into two types: \textit{integrated models} where the LMM and image generation model are connected via neural modules, and \textit{pipeline models} where the LMM and image generation model are connected via prompts in natural language.
The integrated models include: \textbf{(1) MiniGPT-5~\cite{minigpt5}} which connects a large language model with a stable diffusion model via generative vokens, enabling description-free multimodal generation; 
\textbf{(2) GILL~\cite{gill}} which allows a pretrained large language model to generate multimodal responses by mapping the hidden states of text into the embedding space of an image generation model; 
\textbf{(3) EMU-2~\cite{emu2}} which induces in-context learning capabilities of LLMs by scaling up the model size and the size of the pretraining dataset; 
\textbf{(4) EMU-2 Gen + Gold Text} where EMU-2 Gen is a pretrained EMU-2 model instruction-tuned on various controllable image generation tasks. However, EMU-2 Gen cannot generate text so we combine it with ground-truth textual responses to come up with a complete text-and-image interleaved content for evaluation.
The pipeline models include: 
\textbf{(5) GPT-4o~\cite{OpenAI2024} + DALL·E 3~\cite{BetkerImprovingIG}} where GPT-4o is the state-of-the-art proprietary LMM that can comprehend interleaved text-and-image inputs and generate text-only responses. We leverage GPT-4o to generate text responses as well as captions for image responses in the desired positions. Then the captions are fed into DALL·E 3 to generate images. Finally, we combine the text responses with generated images in their original orders;
\textbf{(6) Gemini-1.5~\cite{gemini} + SDXL~\cite{sdxl}}: we build this baseline in a similar way as GPT-4o + DALL·E 3 but use Gemini-1.5 Pro as the LMM and Stable Diffusion XL Turbo as the image generation model. 

\paragraph{Baseline Metrics}
We adopt the following metrics as baselines to validate the effectiveness of our \eval{}.
\textbf{(1) BERTScore} is a reference-based metric for text evaluation. We apply BERTScore to compute the similarity between the text output and the reference in our dataset. We set the BERTScore to 0 if the text output is empty.
\textbf{(2) CLIPScore} 
is originally a reference-free evaluation metric for image captioning, which computes the cosine similarity between the CLIP embeddings of a predicted caption and that of the input image. We adopt CLIPScore as two baselines: a reference-based metric to compute image-image similarity between predicted images and ground truth images in a pair-wise manner, and a reference-free metric to compute the text-image compatibility between the generated images and text.
\textbf{(3) DreamSim} is a recently proposed model-based metric to measure perceptual similarity. Similar to image-image CLIPScore, we use DreamSim to compute the perceptual distance between predicted images and ground truth images in a pair-wise manner.

\subsection{Main Results}
We show the main results of using \eval{} to conduct the fine-grained evaluation for various baseline approaches on \dataset{} in Table~\ref{tab:main_result}. The baselines in the upper part are the \textit{integrated} and \textit{open-sourced} models while the baselines in the lower part are the pipeline models where the LMMs are proprietary. From Table~\ref{tab:main_result}, we observe that: \textbf{First}, the pipeline models consistently outperform the integrated models on all evaluation aspects by a significant margin, where GPT-4o + DALL·E 3 achieves the best performance on \textit{helpfulness} and the average score of all the aspects. This indicates that building a strong interleaved generation model for general purposes remains a significant challenge. \textbf{Second}, the pipeline models achieve significantly good performance on \textit{text quality} since Gemini and GPT-4o have strong text generation capabilities. Also, the generated visual prompts are generally coherent with the text content and they are directly fed into the image generation model, so the performance on \textit{text-image coherence} of pipeline models is also remarkable. \textbf{Third}, we observe that the common errors of integrated models include the output text and/or images being empty, in poor quality, or having severe duplication. This is probably due to their weak instruction-following abilities. 
\textbf{Fourth}, \textit{image coherence} is the most challenging aspect for the pipeline models. This is because the image generation model cannot take the images in the input context or previously generated images as conditions. Thus, the generated images do not have strong coherence. 

Given the closed-source nature of GPT-4o, the evaluation based on GPT-4o can be less transparent and sometimes may not be fully reproducible. To this end, we also implement our \eval{} using the current state-of-the-art open-sourced LMM, i.e., LLaVA-NeXT-Interleave~\cite{llava_interleave}, which supports interleaved text and image inputs. We report the results in Table~\ref{tab:main_result_llava} in Appendix~\ref{app:addition_result}.

\subsection{Human Evaluation}
In addition to automatic evaluation, we also conduct an extensive human evaluation to benchmark the baselines and also provide a meta-evaluation on our \eval{} and other evaluation metrics by computing the correlation between automatic evaluation scores and human judgments.
\paragraph{Human Evaluation Setup}
We adopt the same fine-grained evaluation criteria as \eval{}, where for each sample, the annotators need to give a score for each aspect defined in Table~\ref{tab:asp_def}. The only difference is that, instead of rating on a scale of \{0, 1, 2, 3, 4, 5\}, we use a scale of \{0, 1, 2, 3\} for each aspect, where 1, 2, and 3 indicate the quality is \textit{bad, fair}, and \textit{good}, respectively. In this way, we can reduce the difficulty of human evaluation and improve its efficiency. Due to the cost of human evaluation, we select four representative baselines to evaluate, i.e., GILL, EMU-2, Gemini1.5 + SDXL, and  GPT-4o + DALL·E 3. We include more details on human evaluation setup in Appendix~\ref{app:human_eval}.

\begin{figure*}[htbp]
	\centering
	\includegraphics[width=\linewidth, trim={0.06cm 0 0.5cm 0},clip ]{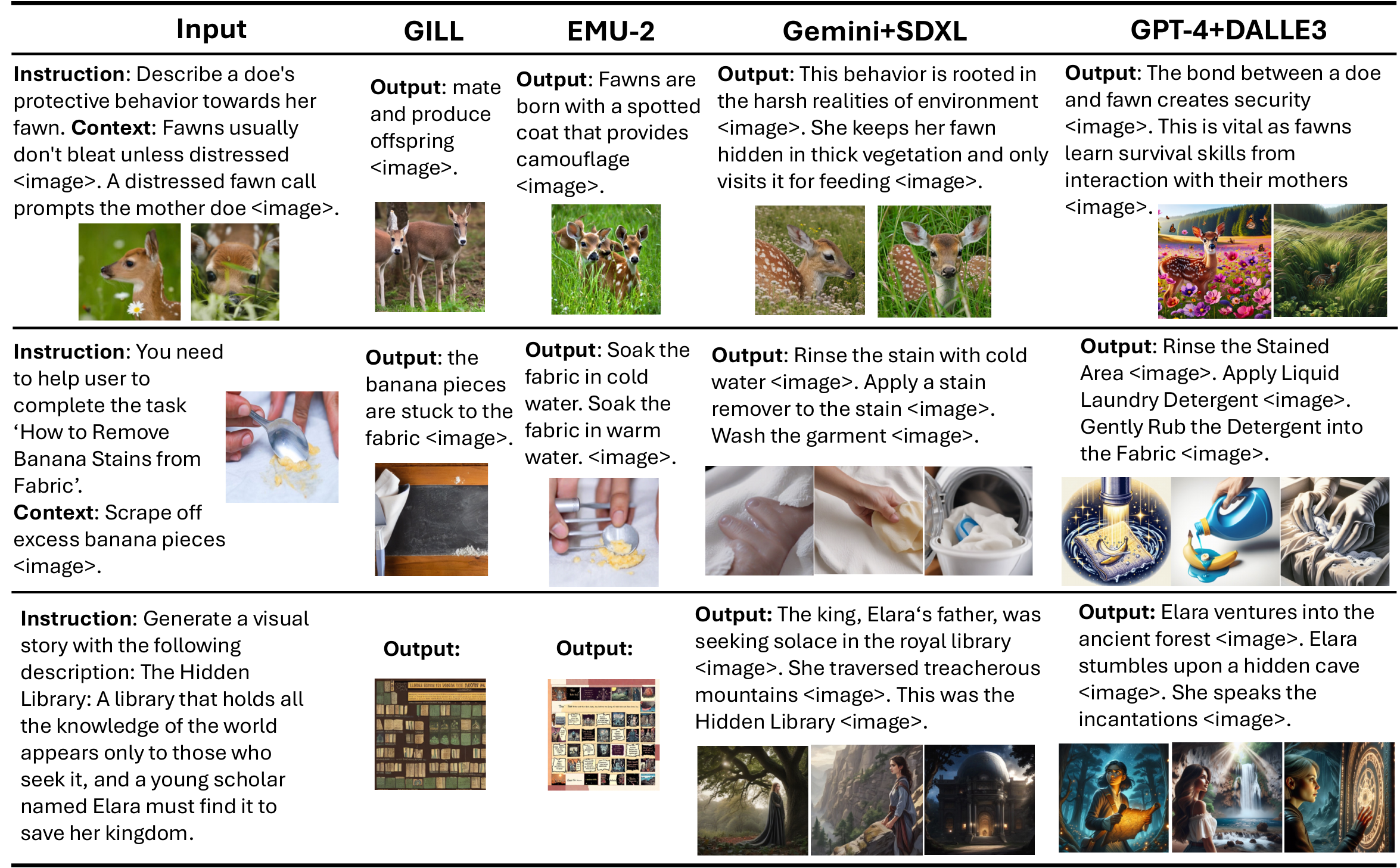}
	\caption{Case study. We select the representative examples of the system outputs from GILL, EMU-2, Gemini+SDXL, and GPT-4+DALLE3.
 }
\label{fig_case_study}
 \vspace{-3mm}
\end{figure*}

\paragraph{Results}
We show the human evaluation results in Table~\ref{tab:human_result}. The human evaluation is generally consistent with the automatic evaluation in Table~\ref{tab:main_result}. The pipeline models consistently outperform integrated models by a large margin, where GPT-4o+DALL·E 3 also achieves the best performance on \textit{helpfulness} and the average performance. There's significant room for improvement in the integrated open-sourced models.
We report the Inter Annotator Agreement (IAA) in Table~\ref{tab:human_iaa} in Appendix~\ref{app:human_eval}.

\begin{figure*}[ht!]
	\centering
	\includegraphics[width=\linewidth, trim={0 0 0 0},clip ]{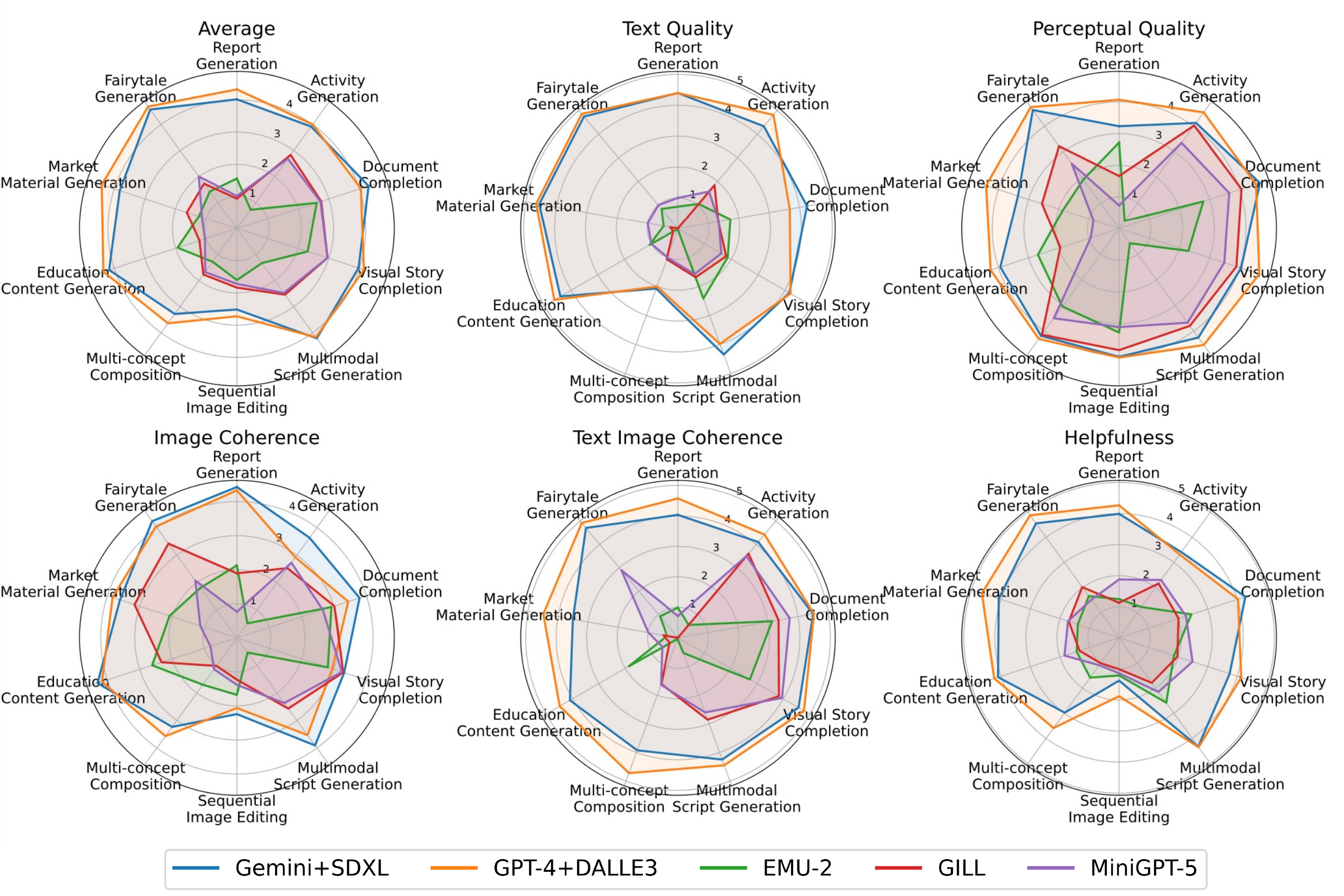}
	\caption{Radar figures of evaluation results on each evaluation aspect for each task.
 }
\label{fig_radar}
 \vspace{-3mm}
\end{figure*}

\paragraph{Correlation Analysis}
To validate the effectiveness of our proposed metric, we conduct a correlation analysis by comparing the evaluation results from automatic metrics with our human evaluation results. Since the baseline metrics only predict an overall score for each instance, we use the same set of evaluation scores to compare against the human rating on each aspect separately. For \eval{}, we compare evaluation scores with the human rating on corresponding aspects. Since most baselines require a gold reference, we use the context-based subset, where each instance has an associated reference output, to compute the correlation. From Table~\ref{tab:meta_eval}, our \eval{} consistently outperforms previous metrics by a significant margin in every aspect. Our metric has a particularly higher correlation on \textit{text quality}, which is because \textit{text quality} is relatively easier to evaluate with large language models like GPT-4o~\cite{zheng2023judging}. Our metric achieves the lowest correlation on \textit{perceptual quality}. The plausible reason is that GPT-4o's perceptual recognition capability is still not strong enough to accurately detect visual artifacts or unnatural disruptions in the images~\cite{blink}.
We also find that baseline metrics generally achieve poorer correlation, e.g., most metrics except for BERTScore almost do not have any correlation with \textit{helpfulness}.
BERTScore achieves the best correlation on \textit{helpfulness} among baseline metrics, which indicates that text quality could be a good indicator of whether the overall interleaved content is helpful.

In addition, we also report the correlation with human judgments of InterleavedEval based on the open-sourced LLaVA-NeXT-Interleave in Table~\ref{tab:meta_eval}. InterleavedEval-LLaVA can achieve promising correlations with humans, generally surpassing previous metrics by a large margin. While there is still a significant gap between the performance of GPT-4o and LLaVA, probably due to the limited capability of LLaVA-NeXT-Interleave, we believe our InterleavedEval-LLaVA is a good alternative to InterleavedEval-GPT-4o in the scenarios where transparency and reproducibility are highly desired. We leave how to build a more powerful open-sourced evaluator for future work.







\section{Discussions}
\label{sec:discussion}

\begin{table*}[ht]
\center 
\resizebox{0.95\textwidth}{!}
{
\begin{tabular}{l|c|c|c|c|c|c}
\toprule
\textbf{Model} & \textbf{Text Quality} & \textbf{Perceptual Quality} & \textbf{Image Coherence} & \textbf{TIC} & \textbf{Helpfulness} & \textbf{AVG} \\ 

\midrule

MiniGPT5 & 1.29 & 3.47 & 2.04 & 2.64 & 1.76 & 2.24 \\
GILL & 1.37 & 3.96 & 2.01 & 2.61 & 1.51 & 2.29 \\
EMU-2 & 1.29 & 2.22 & 1.65 & 1.18 & 1.84 & 1.64 \\
\midrule
Gemini1.5+SDXL & \textbf{3.29} & 4.24 & \textbf{3.26} & 3.94 & 3.25 & 3.60 \\
GPT-4o+DALLE3 & 3.12 & \textbf{4.39} & 3.08 & \textbf{4.36} & \textbf{3.48} & \textbf{3.69} \\
\bottomrule
\end{tabular}
}
\caption{\textbf{Automatic evaluation} results of the \textit{context-based} subset on \dataset{}. TIC is the abbreviation for 'Text-Image Coherence'. The best results are highlighted in \textbf{bold}. }
\label{tab:result_context_based}
\end{table*}

\begin{table*}[ht]
\center 
\resizebox{0.95\textwidth}{!}
{
\begin{tabular}{l|c|c|c|c|c|c}
\toprule
\textbf{Model} & \textbf{Text Quality} & \textbf{Perceptual Quality} & \textbf{Image Coherence} & \textbf{TIC} & \textbf{Helpfulness} & \textbf{AVG} \\ 

\midrule

MiniGPT5 & 1.00 & 1.09 & 1.07 & 1.06 & 1.78 & 1.20 \\
GILL & 0.12 & 2.23 & 2.58 & 0.23 & 1.45 & 1.32 \\
EMU-2 & 0.77 & 2.35 & 2.20 & 1.05 & 1.38 & 1.55 \\
\midrule
Gemini1.5+SDXL & 4.50 & 3.66 & \textbf{4.13} & 3.98 & 4.10 & 4.07 \\
GPT-4o+DALLE3 & \textbf{4.60} & \textbf{4.31} & 4.05 & \textbf{4.52} & \textbf{4.41} & \textbf{4.38} \\
\bottomrule
\end{tabular}
}
\caption{\textbf{Automatic evaluation} results of the \textit{context-free} subset on \dataset{}. TIC is the abbreviation for 'Text-Image Coherence'. The best results are highlighted in \textbf{bold}. }
\label{tab:result_context_free}
\end{table*}

\begin{table*}[!ht]
\center 
\resizebox{0.9\textwidth}{!}
{
\begin{tabular}{l|c|c|c|c|c|c}
\toprule
\textbf{Output Steps} & \textbf{Text Quality} & \textbf{Perceptual Quality} & \textbf{Image Coherence} & \textbf{TIC} & \textbf{Helpfulness} & \textbf{AVG} \\ 

\midrule

Less & 1.8 & 1.1 & 1.2 & 1.3 & 2.1 & 1.5 \\
Equal & \textbf{2.7} & \textbf{3.8} & \textbf{4.0} & \textbf{4.0} & \textbf{3.0} & \textbf{3.5} \\
More & 1.7 & 3.5 & 2.4 & 3.3 & 2.0 & 2.6 \\

\bottomrule
\end{tabular}
}
\caption{Analysis of the number of output steps compared with ground truths.}
\label{tab:impact_output_step}
\vspace{-3mm}
\end{table*}

\paragraph{Qualitative Analysis }
We conduct a qualitative analysis of benchmarked models in Figure~\ref{fig_case_study} and have the following observations: (1) while GILL can generate images with reasonable quality, the generated text and images are typically not coherent with the instruction and context. In the example in the first row, the generated text is totally irrelevant to the task, while the image is also inconsistent with input images. (2) EMU-2 can often generate text that is relevant to the task, but the quality is not good enough. In the example in the second row, it repeatedly says ``soak the fabric in water'' but does not contain other useful content. Another weakness of EMU-2 is its poor conditional image generation capability, where generated images have obvious visual distortions and could be duplicated with input images. (3) On the other hand, the pipeline models can generally better follow the instructions and generate text and images in higher quality. Nevertheless, they still occasionally have some drawbacks. For Gemini+SDXL, some of the generated images (e.g., the first output image in the second example) still have obvious defects. For GPT-4+DALLE3, the style of generated images can be dramatically different from input images, as DALLE3 is prone to generate images in cartoon or dramatic styles. (4) Maintaining image coherence, i.e., the coherence of style and entities across images, is still very challenging for most models. In the third example, for the pipeline models, the same character has a very different appearance across the images, which makes the content inconsistent. (5) For the instances on the context-free subset, the integrated baselines have significantly worse performance, where they only generate one image with extremely poor quality. We hypothesize the reason is that those models cannot truly understand and follow the instructions. 
To sum up, our qualitative analysis indicates there is still significant room for improvement in interleaved generation.

\paragraph{Breakdown Results on Each Use Case}
We show a detailed breakdown of the average results on all the aspects of each use case. From Figure~\ref{fig_radar}, we observe that (1) for pipeline-based models, image editing and subject-driven generation achieve the lowest results, whereas the models can achieve scores above 4 on other use cases; and (2) integrated models typically achieve low performance on the context-free subset in \dataset{}. The potential reason is that these models did not specifically fine-turned on the data with text-only inputs, and thus cannot generate interleaved content well.

\paragraph{Breakdown Performance on Context-based and Context-free Subsets}
We show the breakdown performance on the context-based and context-free subsets of \dataset{} in Table~\ref{tab:result_context_based} and Table~\ref{tab:result_context_free}. Our findings are: (1) pipeline baselines consistently outperform integrated baselines on both subsets; (2) pipeline baselines have better performance on the context-free subset than the context-based subset, while integrated baselines have better performance on the context-based subset than the context-free subset.
Based on the results and our observations, we find the following reasons that could contribute to the discrepancy in performance: (1) pipeline approaches first generate the text along with captions with target images, which can be considered as a planning stage to provide the basis on what images should be generated, making generated interleaved content more useful and reasonable; (2) using separate models (LLMs for text generation and T2I models for image generation) facilitates the generation of high-quality content in each modality; (3) Existing integrated models may struggle with the context-free subset because they haven’t been trained on data with text-only inputs and interleaved multimodal outputs.

\paragraph{Impact of the Number of Output Steps}
We conduct an analysis of how the number of output steps affects the performance when compared with that in the ground truths. 
We calculate the performance of GPT4o-DALLE3 under three cases: the number of predicted steps is less, equal to, or larger than that in the ground truth (“Less”, “Equal”, “More”).
From Table~\ref{tab:impact_output_step}, when the number of predicted steps is less than the ground truths, the model performance is generally worse. This indicates that instances with fewer steps are considered as lower quality and less helpful. When the model has more output steps than ground truths, the performance on text quality, image coherence, and helpfulness are lower. This is because we observed the models produce more images than necessary. Often, these output images are repetitive of the input images or previously generated images. Since we explicitly penalize such repetition in our evaluation criteria, the performance for these cases is lower.

\section{Conclusion}
We introduce \dataset{}, the first benchmark for the evaluation of interleaved text-and-image generation. We also propose \eval{}, a strong multi-aspect reference-free evaluation metric based on GPT-4o. With extensive experiments, we first verify that our proposed metric can achieve significantly higher agreement with humans compared with existing metrics. Through the lens of \eval{}, we then observed that while the pipeline models based on proprietary LMMs consistently outperform open-source models, interleaved generation is still a challenging task that requires further advancement.

\section{Limitation}

While our proposed \dataset{} and \eval{} provide a comprehensive evaluation suite for text-and-image interleaved generation, there are still several limitations in our work that we leave for future research. First, while \eval{} achieves the best correlation with human judgments among other evaluation metrics, it still does not have a high correlation on certain aspects, such as perceptual quality, image coherence, and text-image coherence. To further improve the evaluation accuracy, we may need to improve the capability of foundation multimodal models such that they are capable of recognizing subtle but critical differences. Second, our work did not extensively address the bias in using GPT-4 for evaluation, which we consider an important topic for future research.

\section*{Acknowledgement}
This research is partially supported by a research award from Intuit AI Research, the award No. 2238940 from the Faculty Early Career Development Program (CAREER) of the National Science Foundation (NSF), and the U.S. DARPA ECOLE Program \#HR001122S0052. The views and conclusions contained herein are those of the authors and should not be interpreted as necessarily representing the official policies, either expressed or implied, of the U.S. Government. The U.S. Government is authorized to reproduce and distribute reprints for governmental purposes notwithstanding any copyright annotation therein. 

\bibliography{custom}

\newpage

\appendix

\section{More Details on \dataset{}}
\label{app:dataset}


\paragraph{Data Filtering Pipeline} 
To collect the source data from web resources, we first only keep the samples with 3 to 6 images and less than 12 sentences such that the ratio between text and image is balanced. We then apply Llama-8B-Instruct as a text filter to save the data with good text quality. We also apply LPIPS~\cite{lpips} to discard the instances with duplicate images.
\paragraph{Manual Data Selection}
We apply a manual data selection and instruction annotation process to ensure data quality. We select the instances based on the criteria in Table~\ref{tab:asp_def}. We also encourage the annotators to select diverse instances.
\paragraph{Instruction Annotation}
For each instance, we first ask an annotator to draft an instruction, and then ask another annotator to revise the instruction, until both annotators agree that the instructions are of high quality. The annotators are Ph.D. students with expertise in NLP and multimodal learning areas.

\section{More Details on Evaluation}
\label{app:evaluation}
We present the full list of our defined aspects and their definition in Table~\ref{tab:asp_def}.

\begin{table*}[!ht]
    \centering
    \resizebox{\textwidth}{!}
    {
    \begin{tabular}{l|m{14cm}}
    \toprule
        \textbf{Aspect} & \textbf{Definition} \\ \midrule
        Text Quality & Text quality measures how clear, coherent, and error-free the output text is. It considers grammar, spelling, readability, coherence with the instruction and context, and whether it contains duplicate content. \\
        \midrule
        Perceptual Quality & Perceptual quality measures how visually convincing, natural, and free from distortions or artifacts a generated image appears. It considers how accurately the image mimics reality without unnatural disruptions in structure, colors, or composition. \\
        \midrule
        Image Coherence & Image coherence measures the consistency in style and subject representation across images. This includes textures, color palette, lighting, rendering styles, and maintaining consistent physical attributes, clothing, and behavioral traits. Image coherence also penalizes image duplication, where the output images are too similar, or within the output images themselves. \\
        \midrule
        Text-Image Coherence & Text-to-image coherence measure the alignment and integration between textual and visual elements in a pairwise manner, ensuring they work together to convey a unified and cohesive narrative. \\
        \midrule
        Helpfulness & Helpfulness measures how well the output text and images follow the task instructions and provide complete information to achieve the task. It also considers whether the outputs follow a reasonable logic flow. \\
        
        \bottomrule
    \end{tabular}
    }
    \caption{The full list of evaluation aspects and their corresponding definitions in \eval{}. }
    \label{tab:asp_def}
\end{table*}

\subsection{Human Evaluation}
\label{app:human_eval}
\paragraph{More Details on Human Evaluation Setup}
We sampled 100 instances from \dataset{} as a subset for evaluation and ensure its task distribution is the same as the original distribution. In this way, we have 400 data points where each baseline has inference results on 100 instances. For each data point, we have two different annotators who are Ph.D. or master's students with expertise in NLP or multimodal domains to give ratings independently. 

\paragraph{Inter-Annotator Agreement} We show the IAA of our human evaluation in Table~\ref{tab:human_iaa}. The inter-annotator agreement is reasonably good. Note that the evaluation of interleaved generation is still quite subjective, open-ended, and challenging, even with our carefully designed human evaluation aspects and guidelines.

\begin{table*}[ht]
\center 
\resizebox{0.8\textwidth}{!}
{
\begin{tabular}{c|c|c|c|c|c}
\toprule
\textbf{Text Quality} & \textbf{Perceptual Quality} & \textbf{Image Coherence} & \textbf{TIC} & \textbf{Helpfulness} & \textbf{AVG} \\ 

\midrule

0.689 & 0.606 & 0.620 & 0.627 & 0.619 & 0.612 \\

\bottomrule
\end{tabular}
}
\caption{\textbf{Inter-Annotator Agreement} of human evaluation in terms of Cohen's Kappa score. }
\label{tab:human_iaa}
\end{table*}

\section{Additional Experiment Results}
\label{app:addition_result}
\subsection{Automatic Evaluation Results on based on LLaVA-NeXT-Interleave}

We report the automatic evaluation results in Table~\ref{tab:main_result_llava} based on the existing state-of-the-art open-sourced LMM that supports interleaved text and image inputs, i.e., LLaVA-NeXT-Interleaved~\cite{llava_interleave}, in Table~\ref{tab:main_result_llava}. We use the same evaluation instructions and criteria to prompt the model to predict numerical scores from 1 to 5. We show the automatic evaluation results in Table A and the correlation analysis in Table B. We use the same experiment setup for a fair comparison. 
From Table~\ref{tab:main_result_llava}, the benchmarked performance using InterleavedEval with LlaVA-NeXT-Interleaved generally aligns with human evaluation in Table~\ref{tab:human_result} and automatic evaluation with GPT-4o in Table~\ref{tab:main_result}. For example, the pipeline-based models consistently outperformed the integrated baselines, and GPT-4o-DALLE3 remains the best model overall.

\begin{table*}[t]
\center 
\resizebox{0.95\textwidth}{!}
{
\begin{tabular}{l|c|c|c|c|c|c}
\toprule
\textbf{Model} & \textbf{Text Quality} & \textbf{Perceptual Quality} & \textbf{Image Coherence} & \textbf{TIC} & \textbf{Helpfulness} & \textbf{AVG} \\ 

\midrule

MiniGPT-5 & 2.52 & 2.22 & 2.28 & 1.68 & 2.59 & 2.26 \\
GILL & 1.60 & 3.26 & 3.09 & 1.50 & 3.08 & 2.51 \\
EMU-2 & 2.86 & 2.41 & 2.44 & 1.66 & 3.11 & 2.50 \\
EMU-2 (Gold Text) & 1.44 & 3.31 & 3.30 & 1.51 & 3.25 & 2.56 \\
\midrule
Gemini1.5+SDXL & \textbf{3.70} & 3.86 & 3.79 & 3.73 & 3.78 & 3.77 \\
GPT-4o+DALLE3 & 3.61 & \textbf{4.16} & \textbf{3.93} & \textbf{3.82} & \textbf{3.87} & \textbf{3.88} \\

\bottomrule
\end{tabular}
}
\caption{\textbf{Automatic evaluation} results of existing interleaved generation models on \dataset{} using \eval{} based on open-sourced LLaVA-NeXT-Interleave. TIC is the abbreviation for 'Text-Image Coherence'. The best results are highlighted in \textbf{bold}. }
\label{tab:main_result_llava}
\vspace{-3mm}
\end{table*}


\end{document}